\documentclass{article}
\usepackage{spconf,amsmath,graphicx}
\usepackage{amssymb, bbm, setspace, mathrsfs, color, listings, graphics, multicol, dcolumn, dsfont, subcaption, float, mathabx, booktabs}
\usepackage{caption}
\usepackage{tabularx}
\usepackage{algorithm}
\usepackage[noend]{algpseudocode}
\usepackage{multirow}
\usepackage[hyphens]{url}
\usepackage{soul}

\title{Robust Domain-Free Domain Generalization \\with Class-aware Alignment}
\name{Wenyu Zhang$^{\dagger}$ \quad Mohamed Ragab $^{\dagger\star}$ \quad Ramon Sagarna$^{\dagger}$}
\address{$^{\dagger}$ Institute for Infocomm Research, Singapore \\
    $^{\star}$ Nanyang Technological University, Singapore}

\begin{document}
\ninept
\maketitle
\begin{abstract}
While deep neural networks demonstrate state-of-the-art performance on a variety of learning tasks, their performance relies on the assumption that train and test distributions are the same, which may not hold in real-world applications. Domain generalization addresses this issue by employing multiple source domains to build robust models that can generalize to unseen target domains subject to shifts in data distribution.
In this paper, we propose \textbf{D}omain-\textbf{F}ree \textbf{D}omain \textbf{G}eneralization (DFDG), a model-agnostic method to achieve better generalization performance on the unseen test domain without the need for source domain labels. DFDG uses novel strategies to learn domain-invariant class-discriminative features. It aligns class relationships of samples through class-conditional soft labels, and uses saliency maps, traditionally developed for post-hoc analysis of image classification networks, to remove superficial observations from training inputs. DFDG obtains competitive performance on both time series sensor and image classification public datasets. 
\end{abstract}

\section{INTRODUCTION}
\label{sec:introduction}

While deep neural networks have widely acclaimed classification performance in diverse fields, their performance is often reliant on the assumption that all the train and test examples are sampled from the same distribution.
In practical applications, though, changes in the data generation regimes are common during operation, effectively causing a shift in the data distribution that can significantly deteriorate model performance~\cite{jiang2017intelligent,stisen2015smart,gideon2019speech}. This not only compromises the accuracy of the existing deep model, but also the applicability if there are no available samples from the new distribution.
In response to such issues, \emph{domain adaptation} algorithms \cite{BenDavid2010domainadaptation} aim at learning a model aligning the source distribution(s) and the new target distribution through a common feature representation space. 
For training, these methods require a sufficient amount of source samples and target inputs, and in some cases additional target labels.
In the more extreme field of \emph{domain generalization} \cite{motiian2017unified,huang2020selfchallenging,balaji2018metareg,albuquerque2020distmatch}, no target samples are available at train time. 
The goal is to learn a robust model that can directly generalize when unseen new data distributions arise. This allows to address many realistic scenarios in which the model needs to be deployed in the wild and new data cannot be collected for constant retraining.  
For example, in object recognition, new images of the same object with different background types may be encountered~\cite{huang2020selfchallenging, Matsuura2020DomainGU}; in fault diagnosis from sensing devices, new operating conditions may display unseen signals \cite{zheng2020diagnosis}; or, in speech recognition, an arbitrary number of unknown speakers may emerge~\cite{gideon2019speech}. 

A core aim of domain generalization is to find a common representation space of the multiple source domains such that the class-discriminative information is captured and the domain-specific statistics are ignored~\cite{albuquerque2020distmatch,balaji2018metareg,Mancini2018BestSF,Li2019EpisodicTF}. 
A popular approach is to minimize the discrepancy among the marginal distributions of input features \cite{ li2019adversarial} or the class-conditional distributions \cite{dou2019semantic}. 
One important limitation is the need for domain labels to identify the groupings of samples for alignment~\cite{Matsuura2020DomainGU}. Typically in domain adaptation and generalization literature, dataset labels are used as domain labels.
This restrains the applicability of the methods since domain labels may be unavailable in practice, and dataset labels cannot replace domain labels in cases where samples of a dataset are drawn from a mixture of domains, take for example photos taken with different devices on social media websites if device type is the domain.

\begin{figure}[t]
    \centering
    \includegraphics[width=0.9\linewidth]{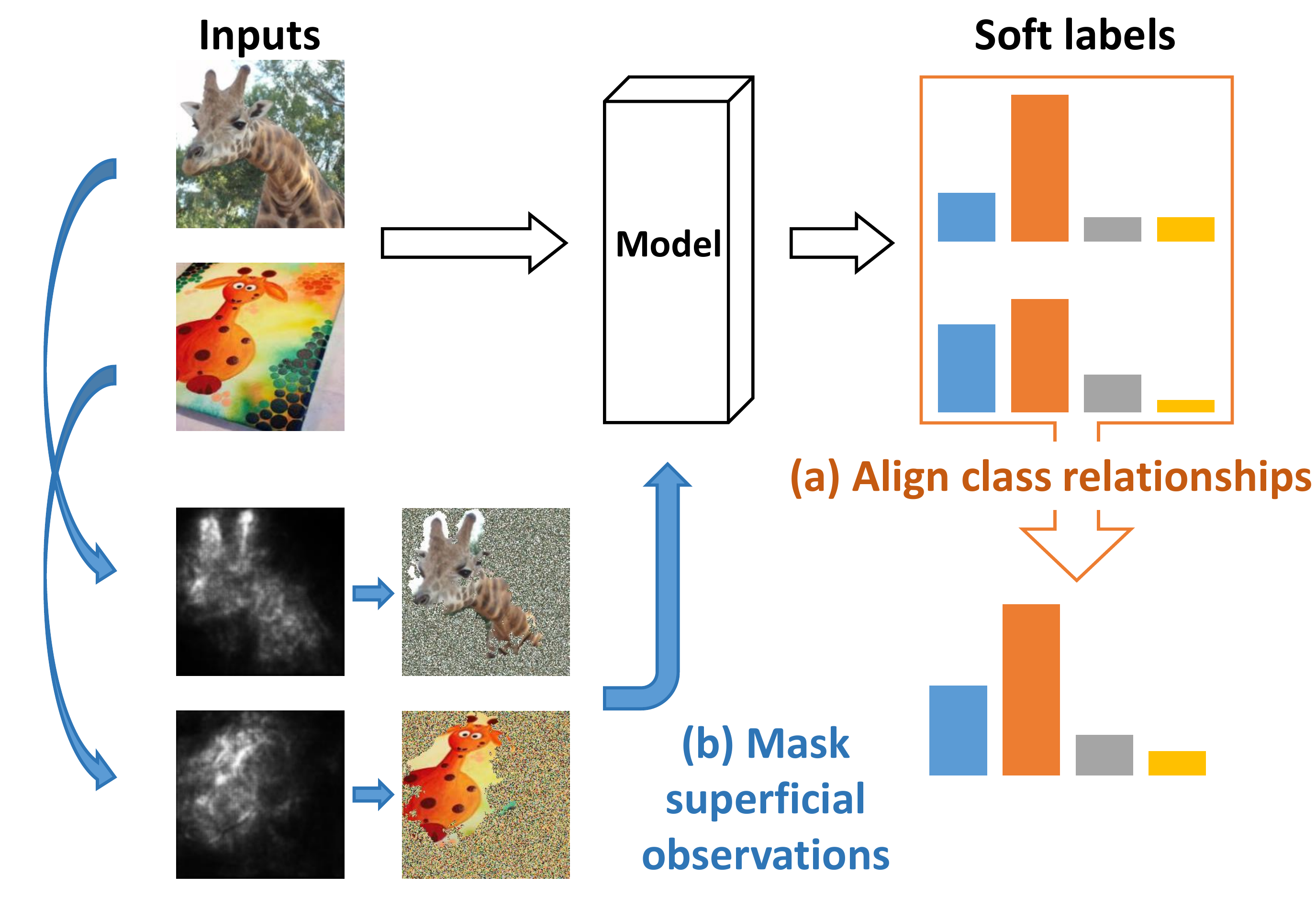}
    \caption{Overview: DFDG learns domain-invariant class-discriminative features without the use of domain labels by (a) aligning class relationships of samples in the same class to the class centroid, and (b) masking superficial observations (e.g. domain-specific backgrounds) from training inputs guided by saliency maps.}
    \label{fig:overview}
\end{figure}

Aiming at a widely applicable domain generalization problem, in this work we propose a \textbf{D}omain-\textbf{F}ree \textbf{D}omain \textbf{G}eneralization (DFDG) method that learns domain-invariant class-discriminative features without the need for domain labels. As illustrated in Figure~\ref{fig:overview}, we propose novel strategies for:
\begin{itemize}
    \item Model regularization to learn domain-invariant features without domain labels by aligning soft labels conditioned on the classes to the class centroid;
    \item Automatic augmentation of inputs to remove superficial, less class-discriminative observations through saliency maps.
\end{itemize}
DFDG is model-agnostic and does not require expert knowledge for feature engineering,  which renders it suitable for learning tasks in general. In our experiments, we consider both time series sensor and image classification benchmarks, and demonstrate that DFDG offers competitive performance compared to state-of-the-art methods. Although saliency maps are developed and conventionally used for image-based applications~\cite{Smilkov2017SmoothGradRN}, our results show empirically that the technique can also be applied on time series sensor data.

\section{RELATED WORKS}
\label{sec:related_works}

\subsection{Combining source models}

With multiple source domains, some works predict for new domains by training a model for each source and combining the respective predictions~\cite{Mancini2018BestSF, vinyals2016matching, innocente2019domain}.
Similarly, some meta-learning frameworks select models for unseen target datasets by exhaustively training all candidate models on each source dataset, and select the final model by matching features and statistics between source and target datasets~\cite{talagala2018metalearn, lemke2010metalearn}. These methods require storing a model for each source domain. In DFDG, only a single model is trained and stored.

\subsection{Learning domain-invariant features}

Many previous works for domain generalization focus on the goal of learning domain-invariant features which are class discriminative. This can be achieved through auxiliary losses with known domain labels. Besides the classification loss, the objective function also includes a domain alignment loss to reduce the distance between samples from different domains~\cite{albuquerque2020distmatch}. This has been implemented through meta-learning~\cite{balaji2018metareg, dou2019semantic, Li2018LearningTG}, or directly minimizing the distance between feature representations of different domains~\cite{motiian2017unified, li2019adversarial}. Some works have an additional separation loss to maximize the distance between samples from different classes to encourage local clustering~\cite{motiian2017unified, dou2019semantic}. In the case where domain labels are unknown, \cite{Matsuura2020DomainGU} shows that samples can be clustered into latent domains and then aligned. Knowledge of accurate domain labels cannot always be assumed in real-world applications, hence we work in the most general setting where domain labels are unavailable.

\subsection{Learning robust features}

Learning features that aim to improve model performance in general by being robust to sample variations at training are found to improve generalization performance in new domains.
Popular approaches to learn robust features is to perturb the network~\cite{Li2019EpisodicTF} or representations during training. Domain-specific representations are removed either from the latent feature space~\cite{Wang2019LearningRR} or from the observed input space~\cite{somavarapu2020stylization}, but these techniques are specific to image inputs. \cite{huang2020selfchallenging} zeros out penultimate-layer feature representations associated with the highest gradient in the final classification layer so that the network learns other features for classification. \cite{fabio2019jigsaw} regularizes the network through an auxiliary self-supervised task to learn spatial correlations within an image.

Other works propose to extrapolate to unseen target distributions by generating adversarial samples~\cite{volpi2018adversarial} or by enforcing equality in training risks across domains ~\cite{Krueger2020OutofDistributionGV}. The latter implies that the worst-performing domain may be improved at the cost of other domains.

\section{Problem setup}
\label{sec:probelm_setting}

We describe the problem setup and notations used in this paper. We denote the set of $K$ source domains as $\mathcal{D}_S = \{D_1, \dots, D_K\}$ and the unseen target domain as $D_t$. For each domain $D$, samples are drawn from a fixed distribution $\left(X^{(D)}, Y^{(D)}\right) \sim P^{(D)}$ where $X$ and $Y$ are the predictor and response variables. We denote the observed samples as $\left\lbrace \left(\boldsymbol{x}^{(D)}_i, \boldsymbol{y}^{(D)}_i \right) \right\rbrace_{i=1}^{N_D}$ for a sample size of $N_D$. We focus on classification tasks in this paper, so $\boldsymbol{y}_i$ is a one-hot vector of the true class label in $C$ classes. No sample from the target domain $D_t$ is available at the training and validation stages of model learning. In the domain-free setting, we further assume that domain labels are unavailable, which is a realistic assumption for data collection. The aim is to train a model $f$ parameterized by $\theta$ to achieve good classification performance on the target domain $D_t$.

A common strategy for classification is to estimate the class probabilities for an input $\boldsymbol{x}_i$. Let $\boldsymbol{p}_i$ be the soft labels or vector of class probabilities for input $\boldsymbol{x}_i$ as learned by the model $f$, that is,
$\boldsymbol{p}_i = \text{softmax}(f(\boldsymbol{x}_i;\theta))$ where $\text{softmax}(\boldsymbol{a})[c] =  \frac{\exp{(a_c)}}{\sum_{j=1}^C \exp{(a_j)}}$
with $a_c$
denoting the element of vector $\boldsymbol{a}$ corresponding to class $c$. The predicted class is the class with the highest probability in $\boldsymbol{p}_i$.

\section{PROPOSED METHOD: DFDG}
\label{sec:proposed_method}

A key concept in domain generalization is to learn representations which are domain-invariant but class-discriminative. To train the model for classification, we use the cross-entropy loss as the base objective for each batch of samples with indices $B$:

\begin{equation}
    \ell_{ce} = - \frac{1}{|B|} \sum_{i\in B} \boldsymbol{y}_i^T\log(\boldsymbol{p}_{i}).
\end{equation}
where $\boldsymbol{y}_i$ is the one-hot vector of the true label and $\boldsymbol{p}_i$ is the soft label estimated for sample $i$. 
Additionally, DFDG learns soft labels $\boldsymbol{p_i}$'s such that they are aligned across source domains. We combine this with input augmentation so that model training focuses on class-discriminative inputs.

\subsection{Regularizing for class relationship alignment}

Constraining the model by aligning features from different domains tends to learn features that are more robust and applicable across domains, and less domain-specific. Being overly reliant on domain-specific features risks overfitting to the source domains and hence poor performance on out-of-distribution samples where these features are absent. Instead of aligning intermediate features, DFDG aligns the estimated soft labels which are lower dimensional and reflects relationships between class concepts. Alignments in the label space have been applied in domain adaption and domain generalization problems where domain labels are known~\cite{dou2019semantic, kang2019contrastive, cicek2019unsupervised}. Without domain labels, we propose an alignment loss:
\begin{align}
    \ell_{align} &= \sum_{c=1}^C\frac{1}{|B(c)|} \sum_{i\in B(c)} \|\boldsymbol{p}_i - \mu(c)\|_2^2\\    
    \mu(c) &= \frac{1}{|B(c)|} \sum_{i \in B(c)} \boldsymbol{p}_i   
\end{align}
where for each class $c$, $B(c) = \{i| i\in B,  y_i = c\}$ are indices of samples, $\mu(c)$ is the cluster center for the corresponding class, and 
 $\boldsymbol{p}_i$ is the vector of soft labels for sample $i$.
The regularization encourages samples from the same class to have similar class relationships regardless of their domains. The complete objective function with this alignment regularization is
\begin{equation}
    L = \ell_{ce} + \alpha \ell_{align}
\end{equation}
with hyperparameter $\alpha$ balancing the effects from the two parts.

\subsection{Masking superficial observations}

The presence of superficial or domain-specific observations in the input space can lead to overfitting to training data if the model learns to rely on these observations for the classification task. We use the term `observation' here to denote a pixel in an image and a reading at one time step for a time series as examples. In images, these superficial observations can be pixels for backgrounds or styles. We propose augmenting $m\%$ of the samples in each batch at training to mask superficial observations by perturbing the observations least relevant to the desired classification task. We perturb by randomly shuffling the least relevant observations, hence preserving the overall statistics of the input. The intuition is that the trained model will be more robust to variations in these superficial observations in the target domain.

While a binary map that exactly pinpoints the location of class-discriminative features in the input is the most desirable and has been found to improve classification performance~\cite{li2017activity}, such accurate maps require human-in-the-loop processing and cannot be readily available for most datasets. Our strategy also differs from feature muting~\cite{schiller2019relevance} that sets a fixed number of features to zero, which can drastically change sample statistics and risk muting useful features through the fixed threshold. 
In DFDG, we use SmoothGrad saliency maps~\cite{Smilkov2017SmoothGradRN} to rank the relevance of each observation. Saliency maps are developed to explain decisions made by neural network models for image classification by assigning a saliency score to each pixel~\cite{Smilkov2017SmoothGradRN, selvaraju2017gradcam}. A `vanilla' gradient-based saliency map $g(\boldsymbol{x},c)$ of an input image $\boldsymbol{x}$ for class $c$ is obtained by differentiating the class logit of model $f$ with respect to $\boldsymbol{x}$ and taking the square element-wise:
\begin{equation}
    g(\boldsymbol{x},c) = \left(\frac{\partial f(\boldsymbol{x}, \theta)[c]}{\partial \boldsymbol{x}}\right)^{\circ 2}
\end{equation}
where each saliency score reflects the amount of influence that a small change in a pixel has on the class logit. SmoothGrad denoises the saliency map by averaging $n$ replicates of $g(\boldsymbol{x},c)$ where Gaussian noise $N(0,\sigma^2)$ is added to $\boldsymbol{x}$ in each replicate, and we denote the resulting SmoothGrad saliency map as $sg(\boldsymbol{x},c)$. We fix $n=25$ and $\sigma=0.15$, the default in \cite{saliencyrepo}, throughout our experiments. We apply the technique on time series as well as image data.

We rank how class-discriminative an observation is by its saliency score. Observations below a threshold is masked and shuffled. The threshold is the $q^{th}$ percentile of the saliency scores of a image, uniformly sampled from $[0, qMax]$ where $qMax$ is the maximum threshold percentile. In this sampling scheme, less relevant observations will be masked more often and the model will be less reliant on them. Figure~\ref{fig:overview} shows examples of saliency maps and masked input images for $q=70$.

\section{EXPERIMENTS}
\label{sec:experiments}

We evaluate DFDG on 3 public datasets: Bearings fault classification from vibration sensor signals, HHAR heterogeneous human activity recognition from motion sensor signals and PACS image classification. We report classification accuracy by treating each domain in turn as the unseen target domain and the rest as source domains. Details of the datasets and implementation configurations can be found in Section~\ref{sec:dataset}. For all datasets, we compare with TrainAll which is a vanilla training of source samples, RSC~\cite{huang2020selfchallenging} which is a robustness method using feature muting that has achieved state-of-the art performance for image classification including PACS, and MMLD~\cite{Matsuura2020DomainGU} which clusters samples into latent domains using convolutional feature statistics and aligns the domains with a discriminator network.

\subsection{Datasets and implementation}
\label{sec:dataset}

\textbf{Bearings Dataset:} 
The Case Western Reserve University (CWRU\footnote{\url{https://csegroups.case.edu/bearingdatacenter/pages/welcome-case-western-reserve-university-bearing-data-center-website}}) Bearings dataset is widely used for rolling element bearings diagnostics in rotating machines~\cite{smith2015rolling}. Vibration signals are collected with 12kHz sampling rate under 8 different working conditions or domains~\cite{smith2015rolling}. The  operating conditions comprise 4 loading torques and two different locations (drive end, fan end). The system has three major types of faults, namely inner-race fault (IF), outer-race fault (OF), and ball fault (BF), and each fault type occurs with three different dimensions 0.007, 0.014, 0.021 inches. Overall, we have 10 classes with 9 faulty classes and 1 healthy class. 
We extracted samples of length 4096 by applying a sliding window with shift size 290~\cite{zhang2017new}.
We use a 6-layer convolutional neural network (CNN) as feature extractor and a 3-layer fully-connected network as classifier.

\textbf{HHAR:}
The HHAR dataset\footnote{\url{http://archive.ics.uci.edu/ml/datasets/heterogeneity+activity+recognition}} consists of sensor readings to classify six activities, namely Biking, Standing, Sitting, Walking, Stair down, and Stair up. 
Gyroscope and accelerometer motion sensors in 8 smartphones and 4 smartwatches are used for monitoring. 
Nine different users execute these activities without predefined order while carrying the devices. 
We treat each user as a domain and, following a recent work on domain adaption~\cite{garrett2020sensor}, we focus on data from 8 smartphones with accelerometer sensors, pre-processed as in that work. 
We use a 3-layer CNN as feature extractor and a 1-layer fully-connected network for classification~\cite{liu2016adaptive}.

\textbf{PACS:} PACS is an image classification dataset\footnote{ \url{http://www.eecs.qmul.ac.uk/~dl307/project_iccv2017}} containing images from 4 styles or domains, namely Photo, Art painting, Cartoon and Sketch. Each domain has a total of 7 classes: dog, elephant, giraffe, guitar, horse, house and person. Following \cite{huang2020selfchallenging, Matsuura2020DomainGU, fabio2019jigsaw}, we use a ResNet18 network architecture and apply the same image augmentations to the samples, namely random cropping to 80-100\% size, horizontal flipping, color jittering and conversion to grayscale.

We fix DFDG hyperparameters $\alpha=0.1$, $m=50$ and $qMax=70$, with further ablation studies in Section~\ref{sec:ablation}. During training, the alignment regularization and input masking strategies are randomly alternated across batches of size 128. The time series sensor datasets are trained for 2000 iterations, and PACS is trained for 3000 iterations. The learning rate is $0.001$, and decayed by factor $0.1$ after $80\%$ iterations. 
To prevent classes from being unrepresented in batches, minority classes are upsampled uniformly at random, keeping a minority-to-majority class ratio of at least $0.5$ in each batch.
All the results are averaged over runs from 3 different seeds.
RSC\footnote{\url{https://github.com/DeLightCMU/RSC}} and MMLD\footnote{\url{https://github.com/mil-tokyo/dg_mmld}} were run using their open-source implementations on the same seeds.

\subsection{Generalization performance}

\begin{figure}
\centering
  \begin{subfigure}[b]{0.49\linewidth}
    \includegraphics[width=\textwidth]{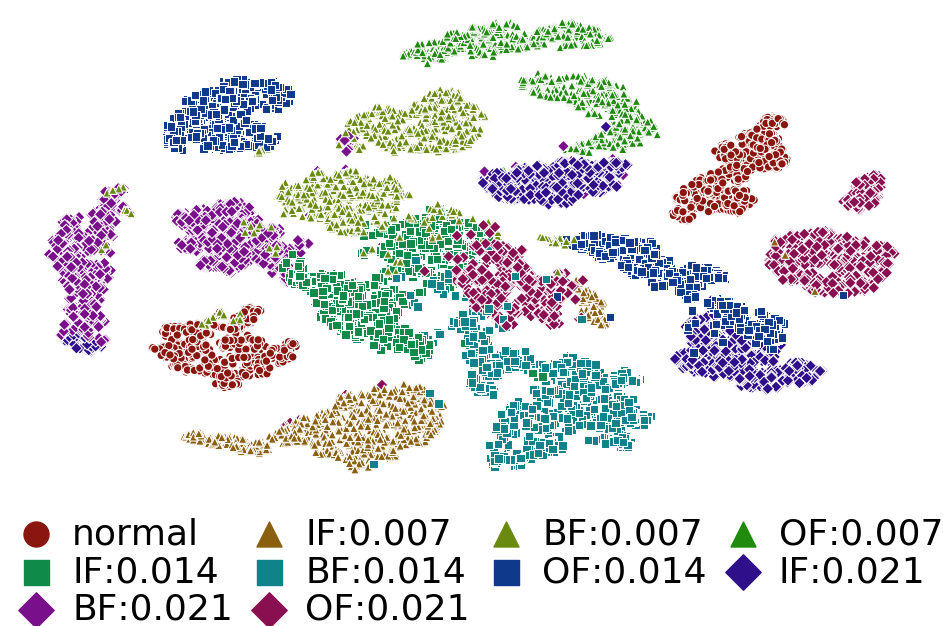}
    \caption{Marked by class}
    \label{fig:tsne_class}
  \end{subfigure}
  \begin{subfigure}[b]{0.49\linewidth}
    \includegraphics[width=\textwidth]{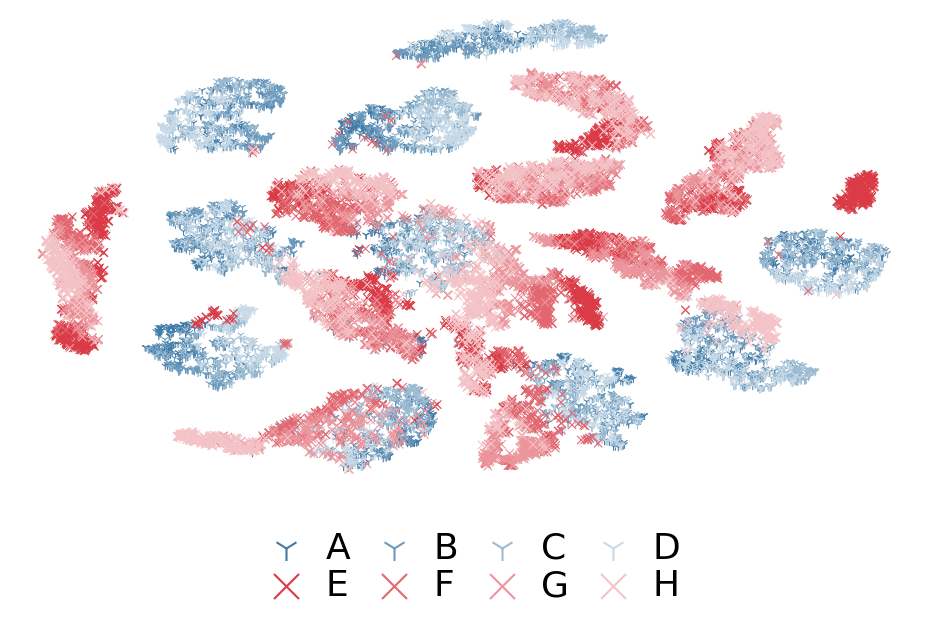}
    \caption{Marked by domain}
    \label{fig:tsne_domain}
  \end{subfigure}
  \caption{t-SNE of Bearings features from DFDG trained with target domain H, demonstrating per-class clustering across domains.}\label{fig:tsne}
\end{figure}

Table~\ref{tab:bearings_dg} shows the classification performance on unseen target domains for the Bearings dataset. Since time series may be less distinguishable across domains than images, clustering with convolutional feature statistics as originated from image stylization~\cite{Matsuura2020DomainGU} may have resulted in incorrect latent domains and hence negative transfer in MMLD. Our proposed DFDG improves average classification accuracy by 2.83\% over the baseline TrainAll. Combining DFDG with RSC, by randomly alternating between the two strategies across batches at training, attains the best accuracy of $84.65\%$.
This shows that DFDG can complement other domain generalization methods.
Figure~\ref{fig:tsne} plots the 2-dimensional t-SNE representations of DFDG features. From Figure~\ref{fig:tsne_class}, samples from the same class tend to cluster together. 
For exceptions such as the class ``BF: 0.021" where there are two clusters, observe from the corresponding Figure~\ref{fig:tsne_domain} that one cluster correspond to domains (A, B, C, D) with drive-end bearings location, and the other correspond to domains (E, F, G, H) with fan-end bearings location. Samples from the same location are clustered closer together due to greater inter-domain similarity.

\begin{table}[h]
\centering

{\setlength\doublerulesep{0.4pt}   
\begin{tabular}{llllll}
\toprule[1pt]\midrule[0.3pt]
\textbf{Target}      & \multicolumn{5}{c}{\textbf{Accuracy ($\%$)}}                                \\ \cmidrule{2-6}
            & TrainAll  & RSC       & MMLD      & DFDG          & DFDG+RSC  \\ \midrule
A           & 52.33     & 68.43     & 65.70     & 66.60         & 68.23 \\
B           & 90.50     & 91.10     & 95.37     & 89.30         & 90.90 \\
C           & 92.90     & 97.60     & 89.33     & 89.67         & 92.47 \\
D           & 74.97     & 77.73     & 65.87     & 77.73         & 77.97 \\
E           & 70.73     & 70.33     & 64.20     & 73.40         & 74.77 \\
F           & 88.53     & 86.40     & 79.03     & 91.50         & 90.23 \\
G           & 87.20     & 90.20     & 86.83     & 91.40         & 94.30 \\
H           & 85.20     & 90.50     & 75.53     & 85.30         & 88.37 \\
\midrule    
Avg         & 80.30     & 84.04     & 77.73     & 83.13         & \textbf{84.65}\\
\midrule[0.3pt]\bottomrule[1pt]
\end{tabular}
}  
\caption{Bearings: Classification accuracy for 10 fault types \label{tab:bearings_dg}}
\vspace{-1em}
\end{table}
\begin{table}[h]
\centering

{\setlength\doublerulesep{0.4pt}   
\begin{tabular}{lllll}
\toprule[1pt]\midrule[0.3pt]
\textbf{Target}      & \multicolumn{4}{c}{\textbf{Accuracy ($\%$)}}                                \\ \cmidrule{2-5}
            & TrainAll  & RSC       & MMLD      & DFDG  \\ \midrule
A           & 43.27     & 41.28     & 45.82     & 36.76  \\
B           & 48.91     & 44.54     & 60.43     & 70.91 \\
C           & 49.15     & 47.68     & 46.73     & 54.63 \\
D           & 45.73     & 52,23     & 49.93     & 62.60 \\
E           & 46.59     & 44.80     & 52.88     & 66.40 \\
F           & 41.98     & 43.02     & 46.95     & 69.28 \\
G           & 30.36     & 29.99     & 43.90     & 60.64 \\
H           & 54.84     & 57.75     & 52.38     & 47.30 \\
I           & 40.86     & 40.59     & 47.19     & 55.07 \\

\midrule    
Avg         & 45.15     & 45.23     & 49.58     & \textbf{58.18} \\
\midrule[0.3pt]\bottomrule[1pt]
\end{tabular}
}  
\caption{HHAR: Classification accuracy for 6 activities \label{tab:hhar_dg}}

\end{table}
\begin{table}[h]
\centering

{\setlength\doublerulesep{0.4pt}   
\begin{tabular}{lllll}
\toprule[1pt]\midrule[0.3pt]
\textbf{Target}      & \multicolumn{4}{c}{\textbf{Accuracy ($\%$)}}                                \\ \cmidrule{2-5}
            & TrainAll  & RSC       & MMLD      & DFDG  \\ \midrule
Art         & 78.73     & 80.37          & 78.99     & 79.23\\
Cartoon     & 74.30     &  76.84         & 77.06     & 75.84\\
Photo       & 94.55     &   94.99        & 95.41     & 95.45\\
Sketch      & 76.19     &  74.40         & 62.56     & 77.87\\
\midrule    
Avg         & 80.94     & 81.65         & 78.51     &\textbf{ 82.10}\\
\midrule[0.3pt]\bottomrule[1pt]
\end{tabular}
}  
\caption{PACS: Classification accuracy for 7 classes. 
\label{tab:pacs_dg}}

\end{table}

Table~\ref{tab:hhar_dg} shows the classification performance on the 9 unseen target users or domains for the HHAR dataset. Overall, DFDG significantly outperforms the compared approaches by more than 8\%. Additionally, DFDG achieves best accuracy in 7 out of 9 users. 

Table~\ref{tab:pacs_dg} shows the classification performance for the PACS dataset. 
MMLD performed irregularly across our runs on the Sketch domain where images have white backgrounds unlike those in the source domains, and lags clearly behind the other methods. 
DFDG improves performance over the baseline TrainAll uniformly across all 4 domains, and offers the best average accuracy of $82.10\%$ in this comparison.

\subsection{Ablation studies}
\label{sec:ablation}

We perform ablation studies on the effects of alignment regularization and input masking using the Bearings dataset. From Table~\ref{tab:ablation}, we see that regularization and masking both improve classification performance, with the latter having a larger effect. The exact amount of regularization and masking that works best varies by dataset. 
\begin{table}[!h]
\centering

{\setlength\doublerulesep{0.4pt}   
\begin{tabular}{llll}
\toprule[1pt]\midrule[0.3pt]
\textbf{Regularization}     & \multicolumn{2}{c}{\textbf{Masking}}      & \textbf{Avg Accuracy ($\%$)}       \\ \cmidrule{1-3}
$\alpha$                    & m     & qMax                              &   \\ \midrule                    
-                           & -     & -                                 & 79.75 \\
0.01                        & -     & -                                 & 80.21 \\
0.1                         & -     & -                                 & 79.97 \\
-                           & 33    & 30                                & 83.21 \\
-                           & 50    & 70                                & 83.42 \\
\midrule[0.3pt]\bottomrule[1pt]
\end{tabular}
}  
\caption{Ablation study of DFDG on Bearings dataset. Hyperparameter values 0 are represented by '-' (dash). \label{tab:ablation}}

\end{table}

We also study the effect of using different saliency maps for input masking on the Bearings dataset. 
The vanilla saliency map, which takes the derivative of the class logit with respect to the input, results in an average classification accuracy of $80.77\%$. This approach can produce noisy maps giving poor estimates of how class-discriminative an observation is. 
In contrast, SmoothGrad is less noisy, and obtains a noticeably higher accuracy of $83.42\%$.

\section{CONCLUSION}
\label{sec:conclusion}

In this paper, we proposed DFDG, a model-agnostic method for domain generalization that does not require domain labels for training. DFDG aligns class-conditional soft labels and masks superficial observations from training inputs to learn domain-invariant class-discriminative features. 
Overall, DFDG exhibits better performance against competing methods for both time series sensor and image classification problems. Namely, DFDG outperforms TrainAll and the state-of-the-art MMLD and RSC on the PACS image dataset and the HHAR sensor dataset. On the Bearings sensor dataset, RSC obtains a higher accuracy, however we demonstrate that DFDG can complement this method to achieve the best accuracy results.

\bibliographystyle{IEEEbib}
\bibliography{references}

\end{document}